# Ax-to-Grind Urdu: Benchmark Dataset for Urdu Fake News Detection

Sheetal Harris, Jinshuo Liu, Hassan Jalil Hadi, Yue Cao
School of Cyber Science and Engineering, Wuhan University, Wuhan, China.

**Abstract:**
Misinformation can seriously impact society, affecting anything from public opinion to institutional confidence and the political horizon of a state. Fake News (FN) proliferation on online websites and Online Social Networks (OSNs) has increased profusely. Various fact-checking websites include news in English and barely provide information about FN in regional languages. Thus the Urdu FN purveyors cannot be discerned using fact-checking portals. State-of-the-art (SOTA) approaches for Fake News Detection (FND) count upon appropriately labelled and large datasets. FND in regional and resource-constrained languages lags due to the lack of limited-sized datasets and legitimate lexical resources. The previous datasets for Urdu FND are limited-sized, domain-restricted, publicly unavailable and not manually verified where the news is translated from English into Urdu. In this paper, we curate and contribute the first largest publicly available dataset for Urdu FND, "Ax-to-Grind Urdu", to bridge the identified gaps and limitations of existing Urdu datasets in the literature. It constitutes 10,083 fake and real news on fifteen domains collected from leading and authentic Urdu newspapers and news channel websites in Pakistan and India. FN for the Ax-to-Grind dataset is collected from websites and crowd-sourcing. The dataset contains news items in Urdu from the year 2017 to the year 2023. Expert journalists annotated the dataset. We benchmark the dataset with an ensemble model of mBERT, XLNet, and XLM-RoBERTa. The selected models are originally trained on multilingual large corpora. The results of the proposed model are based on performance metrics, F1-score, accuracy, precision, recall and MCC value. F1-score of 0.924, accuracy of 0.956, precision of 0.942, recall of 0.940 and an MCC value of 0.902 demonstrate the effectiveness of the proposed approach for Urdu FND. Comparison analysis with SOTA ML and DL models and existing Urdu benchmark datasets exhibit that the ensemble model outperforms them for Urdu FND. The dataset used for our experiments is publicly available at https://github.com/Sheetal83/Ax-to-Grind-Urdu-Dataset for further analysis and validation.

***Keywords:*** *Ensemble model, Natural Language Processing (NLP), Urdu Corpus, Urdu Fake News Detection (FND)*

**Introduction:**

The epidemic of Fake News (FN) has been witnessed in the recent past. Fake News Detection (FND) is paramount since it has become a global issue in the modern digitalised world. The proliferation of FN on News websites and Online Social Networks (OSNs) during the pandemic (Covid-19)[1] had instigated confusion among the masses. It also caused deaths worldwide since the online platforms encumbered FN in the wake of precautions and fake remedies. FN about Covid-19, such as the origin of the coronavirus is a lab and a biological weapon, gargling bleach and applying alcohol can be helpful, 5G is responsible for its spread, use cocaine to kill Corona virus[2], etc. is just the tip of the iceberg. There are a series of FN and manipulated content that may cause psychological and physical loss and pose a threat to society.

In the airstrikes[3] between India and Pakistan on 26 February 2019, the arch-rivals flooded the OSNs with propaganda and heated comments. Pakistan detained the Indian pilot, which further aggravated the situation. Traditional media also joined the race, which resulted in social upheaval. The war between Russia and Ukraine[4] is another example of the FN explosion on web portals and OSNs. FN on the arrest of US Former President Donald Trump[5], followed by an AI-generated image of the explosion at the Pentagon[6], created confusion among its readers. Such news poses a threat to society. It also comprehends the far-reaching effects of FN worldwide. Thus, FND in the digital era is of paramount importance.

---

[1] https://www.medicalnewstoday.com/articles/coronavirus-myths-explored
[2] https://ranchcreekrecovery.com/blog/does-cocaine-help-with-covid-19/
[3] https://www.stimson.org/2022/three-years-after-balakot-reckoning-with-two-claims-of-victory/

FND is challenging and arduous since it entails multiple factors, such as psychological, financial and political agenda. The malicious intent is indiscernible. Therefore, it is difficult to distinguish between accidentally reported FN and intentionally disseminated FN [1]. The capability of users and human judgement to identify FN is limited. Moreover, their affiliations (social context, religious, political, etc.) are a potential threat that hinders their decision about FN. The echo-chamber effect, psychological and cultural differences and varied perspectives obstruct determining the legitimacy of news content [4]. Evidence suggests that the dynamic propagation of FN compared to real news cannot be negated [3].

Extensive research on FND in a specific language, such as English has been conducted because of the available resources and large datasets [31, 32]. Various fact-checking websites also store and provide factual claims for non-fictional news in English, such as PolitiFact.com[7], FactCheck.org[8] and Snopes.com[9]. These online portals use the technological advantage of digital libraries and evaluate the accuracy and reliability of the factual claims in the content [6]. However, the news is limited in number and does not provide insight into various regional news in different languages. This method remains ineffective and time intensive since general readers lack technological expertise. Therefore, these fact-checked claims cannot correct users' erroneous perceptions of events in Urdu.

According to Ethnologue[10], Urdu is among the "Top 10 most spoken languages, 2023", with 231 million people worldwide. It is widely spoken in Pakistan and India. With limited resources, Urdu FND is still in its infancy. The lack of appropriate language processing tools and benchmark corpora increases the complexity of the FN classification in Urdu.

A growing body of literature recognises the importance of FND in various languages [6, 7]. Regional languages such as Urdu have limited-sized datasets and authentic resources, which is the chief limitation of Urdu FND to date. Amidst the few available datasets, some are in Roman Urdu [4], which cannot be used for Urdu FND. Secondly, the Urdu datasets are generated to accomplish the research goals, such as identifying hate speech and ethnic hatred-related FN [9]. For example, if we require a dataset of reasonable size on political news to address the issue of future information warfare between India and Pakistan. Then, the dataset related to various fields, such as entertainment, sports, showbiz, etc., will not suffice [8]. Hence, the lack of multi-domain and cross-domain datasets is another issue [11]. The existing datasets translated from English news into Urdu show poor performance and are not manually verified [10, 14]. Some datasets are not publicly available for in-depth research [7]. Therefore, to develop and evaluate the SOTA mechanism, a comprehensive FND dataset is required.

Another issue is the availability of automation tools and techniques to validate the accuracy of the shared news in different regional languages on the online portals and OSNs. Natural language processing (NLP) has been used extensively for various text processing-related tasks, such as information retrieval, text generation and classification, FND, named entity recognition, etc. The technique addresses text classification issues [34] in different languages. It converts the plaintext into numerical values based on the language characteristics [33]. NLP concentrates on grammatical constructions, word patterns, term frequency and specific expressions of the input text for classification [35]. Previous research has classified news in different languages using NLP [36, 37].

---

[4] https://time.com/6257372/russia-ukraine-war-disinformation/

[5] https://timesofindia.indiatimes.com/world/us/in-pics-the-day-former-us-president-donald-trump-was-arrested/photostory/99255879.cms

[6] https://news.rthk.hk/rthk/en/component/k2/1701708-20230523.htm

[7] https://www.politifact.com/

[8] https://www.factcheck.org/

[9] https://www.snopes.com/fact-check/

[10] https://www.ethnologue.com/insights/ethnologue200/

The study aims to establish the first largest Urdu dataset, "Ax-to-Grind Urdu", to address the identified research gaps of limited-sized, domain-constrained and publicly unavailable datasets. The dataset contains 10,083 multi-domain Urdu news items from websites of leading newspapers and news channels in Pakistan and India. To maintain the uniqueness of the dataset, we have not included any news translated from English into Urdu. The news items are scraped from the year 2017 to the year 2023. We have selected the websites of leading newspapers and news channels from these countries as Urdu native speakers come from two arch-rival countries. Moreover, FN shared by these countries has a significant and consequential impact on their population with reference to their religious, cultural, social and political diversity. Nevertheless, the far-reaching effects of disseminated FN and propaganda from two nuclear states can significantly influence the overall political canvas of the world.

To provide an automated solution for Urdu FND, the dataset will be evaluated using an ensemble model of mBERT, XLNet, and XLM-RoBERTa. Three pre-trained transformer-based models are selected for the proposed ensemble model. The models are originally trained on large multilingual corpora. The stacking method is used since it enhances the overall performance of the selected models for the assigned task. The results of the proposed ensemble model are compared with the results of each model. We also evaluate different Machine Learning (ML) and Deep Learning (DL) approaches on the developed dataset. Lastly, the performance comparison with the existing literature on Urdu benchmark datasets is executed. The comparison is based on performance metrics, such as F1-score, accuracy, precision, recall and MCC score. In this paper, we bridge the identified gap of the existing research by generating the Ax-to-Grind Urdu dataset. The three-fold contributions of the literature are as follows:

- This literature presents a benchmark Urdu news dataset, "Ax-to-Grind", for FND. It is the first large-scale annotated and publicly available dataset that constitutes 10,083 news items in Urdu. The labelled dataset contains news items on fifteen domains from 2017-2023. Expert journalists annotated the dataset.
- The dataset is curated by collecting the news from online websites of renowned Urdu newspapers and news channels in Pakistan and India since Urdu is a widely spoken language in these two countries. We gathered fake news from different websites and crowd-sourcing.
- We propose a novel ensembling approach to benchmark the dataset. The pre-trained transformer-based models originally trained on different languages are stacked using the MV-V technique. MV-V approach selects the best predicted outcome above a pre-defined threshold. This approach augments the overall prediction performance of the proposed model for the targeted task compared to the result outcomes of individual transformer-based models.
- The performance of the proposed model is analysed and compared to different ML, DL and pre-trained transformer models. The comprehensive analysis of SOTA techniques and performance comparison with the existing Urdu benchmark datasets demonstrate that ensembling is an automated approach for Urdu FND.

The second section reviews the existing literature. The third section details the new dataset "Ax-to-Grind Urdu" and its compilation procedure. The fourth section gleans the methodology for the proposed ensemble model for Urdu FND. The benchmark dataset is evaluated, and the results are presented in the fifth section. The proposed ensemble model is evaluated using the dataset, and results are compared to SOTA ML and DL techniques. Existing Urdu benchmark datasets and their prediction performance is compared to demonstrate the efficacy of the proposed approach using the curated dataset. McNemar's test is used to prove the statistical significance of the acquired results. Finally, the last section concludes the literature.

## 2. Related Work:

Over the last two decades, FND has gained impetus. Various approaches have been followed to address the looming threat of FN. Urdu FND has remained an under-researched area. However, keeping up with the pace of innovation, various ML, DL, and NLP techniques have been used for Urdu FND. The studies that have assessed the efficacy of SOTA techniques for Urdu FND with benchmark datasets have been included in this study.

The authors contributed Bend the Truth dataset containing 900 news articles in Urdu [11]. The dataset covers news from five different domains. However, the dataset has a limited number of samples. For the Urdu FND shared task UrduFake@FIRE2020, 400 news articles were added to the original dataset for testing [28]. Whereas 300 news articles were added to the existing dataset for Urdu FND shared task UrduFake@FIRE2021 [29]. Nevertheless, the dataset does not cover the news from politics, religion, and other social domains, which are sensitive topics and instigate users to share FN on OSNs intentionally and unintentionally. Moreover, the multi-domain data provides an assortment of words that can be used to identify propagation patterns [30]. Therefore, FND techniques applied on a limited-sized and domain constraint dataset cannot be considered perceptible compared to large-sized datasets covering cross-domains and multi-domains because the news propagated in the real-world cover multi-domains [30].

Several studies have used the dataset [11, 28, 29] for Urdu FND using ML, DL, and NLP models. The researchers [2] implemented Term Frequency-Inverse Document Frequency (TF-IDF) and fastText features. They achieved 78.7% accuracy on test data. [12] implemented a combination of pre-training of Robustly Optimized BERT Pre-training Approach (RoBERTa), Character-level Convolutional Neural Networks (charCNN) and label smoothing features to accomplish the targeted task with 90% accuracy. The study [13] used a Bi-directional GRU model for Urdu Fake News Detection and showed an F1-score of 80.78% and an accuracy of 81.8%.

The researchers in [14] used an ensemble technique for Urdu FND. The research carried out the experiments using the dataset by [11] and a new dataset. The new dataset contained 2,000 news, and no domain information was added about the collected news items. Additionally, the new dataset contained English news originally that was translated into Urdu using Google Translate. Furthermore, no manual validation was performed with a limited scope of news from a total of five domains only for the dataset [11]. The results demonstrated a lower accuracy, and the literature could not create an incredibly accurate model.

The authors in the literature [4] established a benchmark dataset on Pakistani News. However, the news is not in the Urdu language. The dataset was employed for FND using various ML techniques, Decision Tree (DT), k-nearest neighbours (KNN), Support Vector Machines (SVM), Logistic Regression and Naive Bayes. Global Vectors for Word Representation (GloVe) and Bidirectional Encoder Representations (BERT) embeddings were used for Long Short-Term Memory (LSTM) and CNN-based DL techniques.

The research work [4] presented a large dataset of Pakistani news. However, the news items were in English. The dataset contained around 11,000 labelled news items. It was evaluated with five ML and two DL techniques. LSTM initialized with GloVe Embeddings that exhibited the best performance, i.e., an F1-score of 0.94. It also proves that a large dataset shows better results and performs better for FND compared to a limited number of samples.

The dataset developed by [5] is of 4,097 Urdu news from multi-domain. However, the news items included in the dataset were mainly local news and were translated from English news. An ensemble of Extra Tree (ET), Random Forest (RF) and Logistic Regression (LR) was used to benchmark the dataset with 93.39% accuracy. The study has not used SOTA DL techniques, which could be implemented widely for Urdu FND to prove the efficacy of the benchmark dataset.

This study proposes a dataset with 10,083 news items from fifteen different domains to address the limitation of previous work in Urdu FND with benchmark datasets. The news is collected from 2017 to 2023. The news items in the dataset cover various domains, such as politics, health, sports, entertainment, technology, weather, agriculture, economy, showbiz, social media, education, women's rights, religion, foreign affairs and international. Multi-domain and cross-domain news were added to the dataset to enrich its lexical diversity, which can also be used to identify propagation patterns of different FN. Moreover, it checks the performance of different SOTA models based on diverse word usage. The experts have manually annotated the dataset, which validates its veracity for FND tasks. It contains regional and international news from renowned Urdu news web portals in India and Pakistan. A few true and fake Urdu news items gathered from leading newspapers and online web sources in Pakistan and India for the dataset are presented in Table I.

Table I: Random News Excerpts from the Raw dataset with the Label, Sources and Domains

| Label | Source | Domain | News Articles |
|---|---|---|---|
| True | Jang Newspaper, Pakistan | Politics | الیکشن آگے بڑھانے کا فیصلہ پارلیمنٹ کو کرنا ہے، فضل الرحمان۔ |
| True | Indian Etemaad, India | Health | سردیوں میں دہلی میں کورونا وائرس کے کیسز روزانہ 15,000 تک بڑھ سکتے ہیں: رپورٹ |
| True | Hindustan Express, India | Foreign Affairs | نئی دہلی، جو اسلام آباد پر ہندوستان کے زیر انتظام کشمیر میں اسلام پسندوں اور علیحدگی پسندوں کی پشت پناہی کا الزام لگاتی ہے، نے اس حملے کا الزام پاکستان پر لگایا۔ |
| Fake | Geo Urdu, Pakistan | Sports | ان کا کہنا تھا کہ میری چوٹ جعلی تھی: حفیظ نے ٹی وی چینلز کے خلاف شکایت درج کرادی۔ |
| Fake | Vishvas News, Pakistan | Religion | تصویر میں پاکستانی اسلامی رہنما کو وہسکی کی بوتل کے ساتھ دکھایا گیا ہے |
| Fake | BBC Urdu, Pakistan | Economy | پاکستانی روپے کے مقابلے میں ڈالر 76 تک گر گیا۔ |

## 3. Ax-to-Grind Dataset

We compile a dataset of 10,083 true and fake Urdu news on fifteen different domains. The news items from various domains are included to evaluate the lexical ability of the pre-trained models, which is not possible in the case of a limited number of news items and news categories. Two fine-grained labels, True and Fake, are assigned to each news item and verified through online web sources. 5,053 fake and 5,030 true news are retained in the dataset to maintain a well-balanced distribution of the news items.

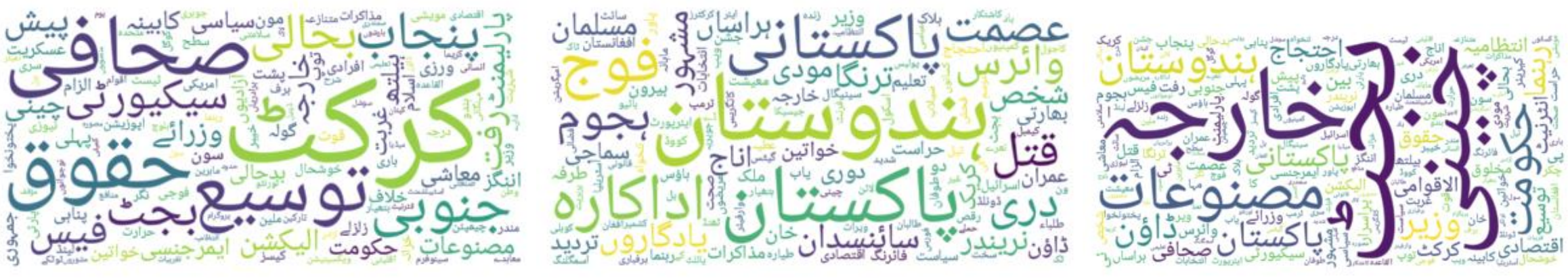

(a) True News (b) Fake News (c) True and Fake News

Figure 1: Word Cloud of True, Fake and Combined True and Fake News from the dataset

Expert journalists manually annotated the dataset. Cohen's Kappa Coefficient [15] is used to determine the agreement rate of the annotated dataset. The statistics results of 0.94 validate its efficacy. Dataset Distribution with respect to Numeric Features is shown in Table II. The word cloud is generated using the Ax-to-Grind dataset. Figure 1. (a), (b), and (c) represents the dataset word cloud of true, fake and both true and fake news (combined dataset), respectively. The unique words and word clouds also demonstrate an overlap of significant words for both true and fake news in the dataset.

Table II: Dataset Distribution with respect to Numeric Features

| Features | True | Fake | Combined |
|---|---|---|---|
| Unique Words | 12,894 | 25,176 | 29,911 |
| Average words per news item | 34.82 | 116.98 | 75.90 |
| Average characters per news item | 171.79 | 528.62 | 350.58 |

### 3.1.1. Dataset Collection and Annotation

There are limited web sources available for the Urdu language. It is the most spoken language in Pakistan and India. Therefore, we have collected authentic news from 21 most popular Urdu newspaper websites and mainstream Urdu news channels in Pakistan. Recently, Geo News11and Dawn12have started fact-checking services. However, the news is in English, and limited news is available since these projects are in a nascent stage.

True news for the dataset is collected from the online websites of Jang, Dawn News Urdu, Express Roaznama, Geo Urdu, BBC Urdu, Indian Etemaad, Inquilab, Hindustan Express, Siasat, etc. The news from the websites of Urdu news channels is also included. We have incorporated Urdu textual content from fifteen categories. Fake news is gathered from different websites, vishvasnews.com13, Sachee Khabar,14 and crowd-sourcing. We have not translated existing datasets from English into Urdu to maintain the originality of the benchmark dataset. We have included captions of fake videos and images shared on OSNs. Professional journalists have cross-checked and validated these captions. The meaningless words and symbols are removed from the scrapped raw data. The sentences are ordered by changing the word sequence. Finally, we have removed incoherent statements and redundant information that affect the model's performance.

### 3.1.2. Corpus Statistics

Table II discloses the number of unique words 29,911words in the dataset. It also shows that FN and True News contain 8,159 common words. We have divided the dataset for training, validation and testing the proposed ensemble model. The split ratio in percentage is 60:20:20 for training, validation and testing, respectively. Table III presents the corpus statistics class-wise.

Table III: The Ax-to-Grind Urdu Dataset Statistics

| | True News | Fake News | Total |
|---|---|---|---|
| Training | 3,018 | 3,031 | 6,049 |
| Validation | 1,006 | 1,011 | 2,017 |
| Testing | 1,006 | 1,011 | 2,017 |
| Total | 5,030 | 5,053 | 10,083 |

### 3.1.3. Dataset Pre-processing

The data is cleaned and pre-processed before it is fed to the models. Therefore, English stop words, URLs, non-alphanumeric characters and special characters are removed. This increases the model performance and reduces the processing time. The following techniques are used:

- Sentence Segmentation: The process is used to separate the input string of a news item from the dataset into relevant sentences. For clarity, a comma (,) known as a "sakta" is put between words and phrases in Urdu. A sentence is completed by adding a full stop (.) called a "khaatma". A question mark (?) is put at the end of an Urdu sentence to indicate a question.
- Stemming: The method changes the words into their base form to lessen input space. It further increases the clarity of the news item by restricting the confusion between the same words. The suffixes of the words are removed from the input words. A few examples of the words from the dataset are given as follows:

| Stem word | Root word | Stem word | Root word |
|---|---|---|---|
| فراہمی | فراہم | لہرائے | لہرا |

- Stop words Removal: Stop words are used frequently in the Urdu language that are conjunctions or words and restrict a sentence grammatically. These words are removed to restrict noise in the dataset because of their limited contextual and semantic importance. An example of removed stop words from the news items is given as under:

| | | | |
|---|---|---|---|
| اور | وہ | یہ | اب |
| میرا | آئی | کا | اس |

- Tokenization: Each news item is tokenized into its respective tokens as pre-trained models can easily interpret tokens. An example of the tokenized news item is given as under:

جموں و کشمیر اننت ناگ میں مسلح تصادم میں لشکر طیبہ کے دو عسکریت پسند مارے گئےجائے وقوعہ پر دھماکے سے 4 دیہاتی زخمی

Token IDs: [5, 214, 18, 96, 90, 8, 167, 68, 154, 70, 7, 154, 9, 18, 77, 71, 18, 18, 53, 51, 63, 18, 77, 9, 18, 77, 71, 18, 9, 18, 90, 24, 75, 10, 156, 77, 20, 11, 69, 82, 7, 28, 77, 94, 18, 12, 150, 71, 72, 24, 77, 4, 32, 72, 71, 18, 72, 63, 7, 18, 56, 72, 63, 18, 10, 82, 90, 63, 18, 6, 10, 7, 52, 74, 63, 18, 76, 24, 82, 86, 18, 71, 11, 72, 69, 77, 70, 82, 18, 70, 7, 49, 82, 63, 76]

## 4. Methodology for Baseline Transformer

We use TF-IDF [16] statistical technique for lexical feature extraction and propose an ensemble of SOTA pre-trained models for Urdu FND.

### 4.1. Lexical Feature Extraction

Frequency, as the name suggests, indicates how frequently a word appears in a news item. It also reflects the importance of the repeated term. Document frequency includes the news items with a particular feature and reveals this frequent feature in the dataset. Hence, the TF-IDF of a specific word rises when repeated in the document and falls with the increase in the number of documents in the corpus that contain the same word.

#### 4.1.2. NLP Pre-trained Transformer-based Models

SOTA NLP pre-trained models are widely used for several languages related tasks. With the advent of Transfer Learning (TL) and pre-trained models, recent breakthroughs in research are feasible. Natural Language Generation and Natural Language Understanding are the two main categories of tasks in NLP. The most recent advancements appear to be fuelled not just by the enormous increases in computing capacity. It is also attributed to different activation functions, fine-tuning and polling strategies. These SOTA methods improve the overall performance of the model. Henceforth, the pre-trained models are authentic to perform various text classification-related tasks, sentiment analysis, text summarization, etc. We have selected three pre-trained transformer-based models for the proposed architecture for Urdu FND to benchmark the dataset. The pre-trained models are selected that are originally trained for multilingual tasks using larger corpora. The pre-trained models mBERT, XLNet, and XLM-RoBERTa [19] will be stacked for an ensemble in this study.

XLNET [17] is an autoregressive pre-trained model. It used the best aspects of Autoregressive (AR) language modelling and Autoencoding (AE). It further confined the drawbacks of AR and AE modelling methods. Furthermore, Transformer-XL, a SOTA AR model, was incorporated into its pre-training. Using an AR technique, the model learns bidirectional contexts and, based on them, predicts the prospective outcomes of the input sequence. mBERT is the multilingual extension of BERT [18], which is trained in 104 languages, including Urdu. XLM-RoBERTa [19] is a multilingual variant of RoBERTa, which is pre-trained in 100 different languages, including Urdu. It is widely used for various language-related tasks in NLP.

#### 4.1.3. Ensembling the Pre-trained Models

The existing literature has widely used the stacking technique for various NLP tasks. Covid Fake Information Detection [20], Named Entity Recognition [21], Fake News Detection [22], and Classification of Thyroid Conditions [23], etc., are some examples. The ensembling technique increases the predicting capability and overall performance of the selected models for an NLP task.

Diversity is the primary criterion for selecting the pre-trained models for an assigned task, as each model is pre-trained on large sized corpus and works differently. The best features of the models are combined with apposite fine-tuning, hyperparameters, activation functions and polling techniques. The overall predictive performance of an ensemble model enhances compared to the individual outcomes of each model. Therefore, we have used the ensembling technique to benchmark the dataset. The proposed model for Urdu FND using the Ax-to-Grind dataset is illustrated in Figure 2.

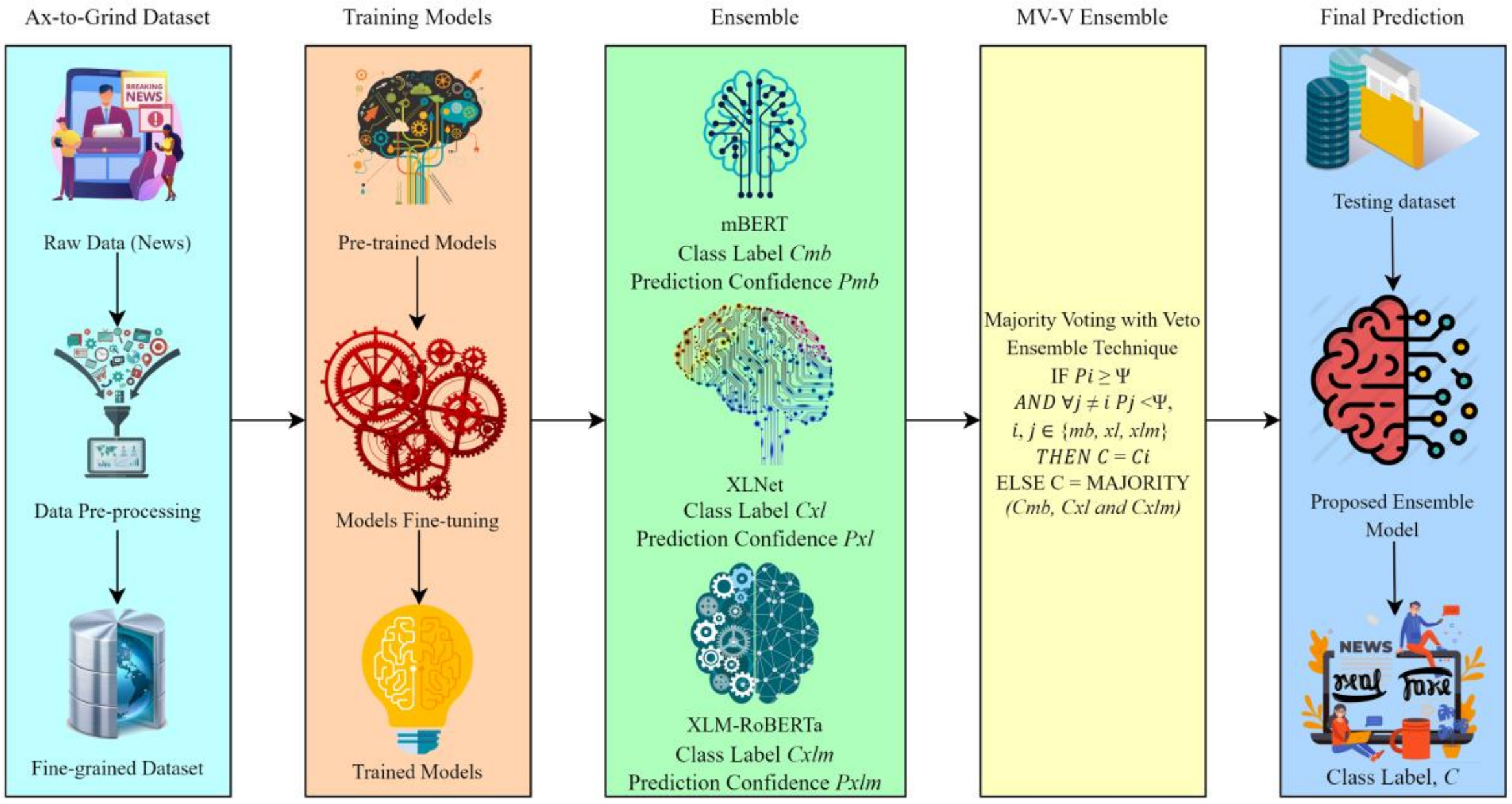

Figure 2: Proposed Architecture for Urdu FND using Ax-to-Grind Dataset

Various stacking and polling techniques were experimented with before the final selected ensembling technique. Support Vector Machines (SVM) and Multi Layer Perceptron (MLP) [24] were used along with polling techniques, Majority Voting (MV) and Majority Voting with Veto capability (MV-V) [25].

An ensemble model acquires the final predictions of the selected models along with their confidence level values. MV retains the final prediction of the selected models. Therefore, the final predicted class of the pre-trained model is selected, which previses with higher confidence. Finally, the predicted class of the model is selected that is above a prediction threshold $\psi$. The stacking method was used since it showed the best performance. Cmb denotes the class label output of mBERT, and Pmb is its prediction confidence. Cxl represents the class label output, and P xl is the prediction confidence of XLNet. Lastly, Cxlm shows the class label output with the prediction confidence Pxlm for XLM-RoBERTa. The following set of rules refers to the MV-V stacking method:

$$IF\ Pi \geq \Psi\ AND\ \forall j \neq i\ Pj < \theta, i, j \in \{\text{mb, xl, xlm}\}$$
$$THEN\ C = Ci, ELSE\ C = MAJORIT\ Y\ (C\text{mb}, C\text{xl}, C\text{xlm})$$

### 4.1.4. Training Phase

The dataset is pre-processed, as explained in Section 3.3. mBERT, XLNet and XLM-RoBERTa are selected for an ensemble. The models are fine-tuned and trained using the tokenized training dataset. We have used the learning rate (2e-5) and AdamW optimizer. Different batch-size is used for the three selected models for the ensemble model. The pre-trained models originally trained for multilingual tasks using larger corpora are selected. We have also used MLP [24] instead of the default Softmax layer at a learning rate of 1e-5. It is adopted to achieve an improved performance of each selected model individually. As we did not ascertain further performance improvements, therefore, we restricted the training phase to 5 epochs.

The performance evaluation validates the selection of the 2x2 hidden layer architecture and the ReLU activation function. The MLP classification heads are trained using the output from the selected models. Limited-memory Broyden-Fletcher-Goldfarb-Shanno (LM-BFGS) solver was used to train the MLP classification head. These choices are based on theoretical and practical considerations. Theoretically, the smaller the dataset, the better the LM-BFGS performs. Secondly, the solver tends to converge faster with the smaller dataset. 5-fold Cross-validation results approve the hyperparameters selection for the ensemble model. Table IV shows the hyperparameters for the selected models.

Table IV: Hyperparameters for each Pre-trained Model

| | Learning Rate | Batch Size | Optimizer |
|---|---|---|---|
| mBERT | 2e-5 | 16 | AdamW |
| XLNet | 2e-5 | 16 | AdamW |
| XLM-RoBERTa | 2e-5 | 32 | AdamW |

### 4.1.5. Testing Phase

The pre-processed and tokenized testing dataset is fed to the trained selected models since the weights have converged. The trained selected models classify the testing dataset. The outputs are introduced to the classification heads. The final class prediction of the testing dataset is acquired. The ensemble model classifies the final class based on the output predictions of the selected models. We have also used Matthews Correlation Coefficient (MCC) [26] as it considers true negative values. Therefore, it presents a better insight into the ensemble model's performance. The performance metrics F1-score, accuracy, precision, recall and MCC determine the efficacy of the proposed ensemble model for Urdu FND.

## 5. Experimental Evaluation

Table V presents the performance evaluation of the ensemble model using 5-fold Cross-validation. mBERT determines an F1-score of 0.896 and an accuracy of 0.902. XLNET shows an F1-score of 0.914 and an accuracy of 0.918. Lastly, XLM-RoBERTa demonstrates an F1-score of 0.902 and an accuracy of 0.912. The proposed ensemble model shows an F1-score of 0.924, accuracy of 0.956, precision of 0.942, recall of 0.940 and an MCC value of 0.902. The results indicate that the stacking method enhances the overall prediction performance of the selected models for Urdu FND. The ensemble model outperforms the efficacy of each selected model for the targeted task.

Table V: Performance Evaluation of the Selected Models for Urdu FND using 5-fold Cross Validation

| **Parameter** | **mBERT** | **XLNET** | **XLM-RoBERTa** | **Ensemble** |
|---|---|---|---|---|
| Accuracy | 0.902 | 0.918 | 0.912 | 0.956 |
| Recall | 0.862 | 0.882 | 0.914 | 0.940 |
| Precision | 0.898 | 0.928 | 0.904 | 0.942 |
| F1-Score | 0.896 | 0.914 | 0.902 | 0.924 |
| MCC | 0.878 | 0.898 | 0.892 | 0.902 |

The comparison with various ML and DL techniques is shown in Table VI. For SOTA ML models, the KNN model shows the best performance on the Ax-to-Grind dataset, followed by SVM. Whereas for DL models, the CNN model outperforms LSTM for Urdu FND. Furthermore, the individual pre-trained transformer-based models perform better compared to ML and DL models. The better performance is demonstrated because the pre-trained models are originally trained on multilingual corpora. The architecture complexity and apposite fine-tuning also enhance their effectiveness for a targeted classification task. Lastly, the ensemble model shows the overall best performance since it comprises the best features of each model. Therefore, it is maintained that the proposed ensemble model is efficacious for Urdu FND.

Table VI: Performance Comparison with ML and DL Models

| **Models** | **F1-Score** | **Accuracy** | **Precision** | **Recall** | **MCC** |
|---|---|---|---|---|---|
| ML Models | | | | | |
| KNN | 0.904 | 0.928 | 0.828 | 0.818 | 0.786 |
| SVM | 0.902 | 0.912 | 0.892 | 0.826 | 0.782 |
| DT | 0.856 | 0.854 | 0.844 | 0.824 | 0.762 |
| RF | 0.834 | 0.818 | 0.764 | 0.748 | 0.724 |
| LR | 0.892 | 0.886 | 0.780 | 0.776 | 0.768 |
| Naïve Bayes | 0.894 | 0.884 | 0.784 | 0.780 | 0.746 |
| Gradient Boost | 0.836 | 0.832 | 0.802 | 0.798 | 0.774 |
| DL Models | | | | | |
| CNN | 0.914 | 0.902 | 0.844 | 0.828 | 0.780 |
| LSTM | 0.892 | 0.886 | 0.798 | 0.792 | 0.702 |

The results of different ensembling techniques are compared and shown in Figure 3. MV-V technique for stacking outperforms MV and SVM. The comparison validates the selection of the MV-V technique for ensembling the results of selected models. It further avoids the overfitting issue for the proposed ensemble model.

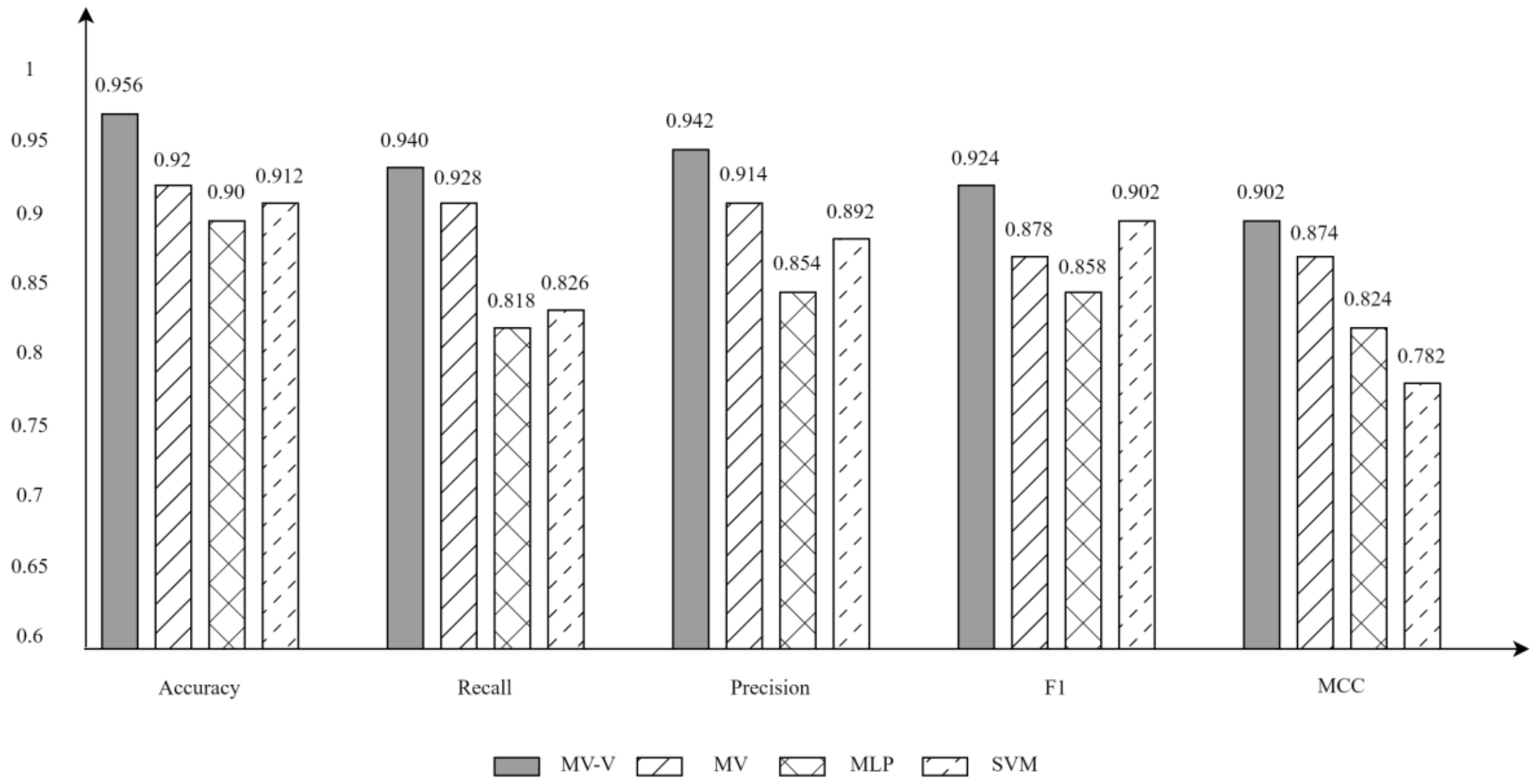

Figure 3: Comparison of various Ensembling Techniques

Table VII illustrates the analytical and performance comparison of the proposed approach for Urdu FND using the Ax-to-Grind dataset with the other Urdu benchmark datasets. The proposed ensemble model shows the best performance using Ax-to-Grind Urdu dataset compared to different approaches used in the existing literature to benchmark Urdu datasets.

Table VII: Analytical and Performance Comparison with the Existing Research with Urdu Benchmark Datasets

| Research | News Items | Domains | F1-Score | Accuracy | Precision | Recall | MCC |
|---|---|---|---|---|---|---|---|
| Bend the Truth [11] | 900 | 5 | 0.90 | 0.88 | - | - | - |
| UrduFake@FIRE2020 [28] | 1300 | 5 | 0.90 | 0.908 | 0.918 | 0.936 | - |
| UrduFake@FIRE2021 [29] | 1600 | 5 | 0.679 | 0.756 | 0.757 | 0.93 | - |
| FND in Urdu Language using ML [5] | 4097 | 9 | 0.932 | 0.938 | 0.889 | 0.963 | 0.862 |
| UFN [14] | 2000 | - | - | 0.833 | - | - | - |
| **Ax-to-Grind Urdu** | **10,083** | **15** | **0.924** | **0.956** | **0.942** | **0.940** | **0.902** |

## 6. McNemar's Test

The statistical significance of the obtained results of the ensemble model is shown using McNemar's Test. The literature [27] has used McNemar's test to evaluate and draw performance comparisons of two different models for FND. The results of McNemar's test are established as 2 x 2 contingency matrices for each selected model and the proposed ensemble model. The results of 5-fold

Cross-validation are used to derive McNemar's test values. Table VIII discloses the values of McNemar's test.

The significance threshold α is established, and the test statistic is calculated. After that, the p-value is calculated, which represents the likelihood of finding the empirical chi-squared value. There exists a statistically significant difference between the models if the p-value is less than the selected significance threshold. The derived values of McNemar's test for the selected models are significant. The p-values of 0.0295 for XLNet, 0.0477 for mBERT and 0.0234 for XLM-RoBERTa support the performance of the proposed ensemble model, i.e., statistically significant compared to each selected model.

| Model | Ensemble Correct ($c_{11}$) | Ensemble Incorrect ($c_{10}$) | Chi-square Value | *p*-value |
|---|---|---|---|---|
| XLNet Correct ($c_{01}$) | 1786 | 22 | 4.738 | 0.0295 |
| XLNet Incorrect ($c_{00}$) | 39 | 155 | | |
| mBERT Correct ($c_{01}$) | 1796 | 32 | 3.920 | 0.0477 |
| mBERT Incorrect ($c_{00}$) | 18 | 156 | | |
| XLM-RoBERTa Correct ($c_{01}$) | 1788 | 54 | 5.069 | 0.0234 |
| XLM-RoBERTa Incorrect ($c_{00}$) | 33 | 127 | | |

Table VIII: Results' Evaluation using McNemar's Test

## 7. Conclusion

FN propagation and dissemination on a larger scale affects its consumers. AI-generated FN has increased the threat intensity that misleads users globally. FND in resource-constrained languages, such as Urdu, has remained an under-researched area due to the scarcity of resources and unavailability of datasets. We followed a three-pronged approach to address the identified limitations in the existing literature. We developed the "Ax-to-Grind Urdu" publicly available dataset. It contains 10,083 fake and true news in Urdu from 2017-2023. The news was amassed from eminent web portals of Pakistani and Indian newspapers and news channels. Second, the included news items incorporate fifteen different domains. Multi-domain and cross-domain news items enhance the dataset lexical diversity to identify FN propagation patterns. Third, we proposed an ensemble model of pre-trained transformer-based models mBERT, XLNet, and XLM-RoBERTa to benchmark the dataset. The prediction performance analysis determines the efficacy of the ensemble model compared to SOTA ML and DL models. Comparison analysis with the existing Urdu benchmark datasets proved the efficacy of the proposed ensemble model. The results exhibited the effectiveness of the proposed approach for Urdu FND. McNemar's test proved that the performance of the ensemble model is statistically significant.

## Acknowledgement

The work is supported by funding source, including the National Key Research and Development Program of China (2020YFA0607902).